\documentclass[sigconf]{acmart}
\AtBeginDocument{%
  \providecommand\BibTeX{{%
    \normalfont B\kern-0.5em{\scshape i\kern-0.25em b}\kern-0.8em\TeX}}}

\setcopyright{acmcopyright}
\copyrightyear{2022}
\acmYear{2022}
\acmDOI{3477495.3531823}

\acmConference[SIGIR '22]{International ACM SIGIR Conference on Research and Development in Information Retrieval}{July 11--15, 2022}{Madrid, Spain}

\acmBooktitle{The 45th International ACM SIGIR Conference on Research and Development in Information Retrieval} 
\acmPrice{15.00}
\acmISBN{978-1-4503-8732-3/22/07}

\usepackage{amsmath}
\usepackage[cal=boondoxo]{mathalfa}
\usepackage{bm}

\usepackage{textcomp}

\usepackage{graphicx}
\usepackage{subcaption}
\usepackage{capt-of}  % allow table in figure
\usepackage{pgfplots}
\usepgfplotslibrary{statistics}
\definecolor{turbo1}{rgb}{0.298 0.573 0.980} % blue
\definecolor{turbo2}{rgb}{0.882 0.290  0.125} % red
\definecolor{turbo3}{rgb}{0.396 0.976  0.467} % green
\definecolor{turbo4}{rgb}{0.949, 0.776, 0.290} % orange
\definecolor{turbo5}{rgb}{0.302, 0.482, 0.925} % darker blue
\usepgfplotslibrary{colorbrewer}
\pgfplotsset{
	compat=1.13,
	legend image code/.code={
		\draw[mark repeat=2,mark phase=2]
		plot coordinates {
			(0cm,0cm)
			(0.1cm,0cm)        %% default is (0.3cm,0cm)
			(0.2cm,0cm)         %% default is (0.6cm,0cm)
		};%
	}
}
\usepackage{array}
\usepackage{multirow}
\usepackage[para, flushleft]{threeparttable}  % para sets inline tablenotes
\makeatletter  % change tablenote size of threeparttable
\g@addto@macro\TPT@defaults{\footnotesize} 
\makeatother
\newcolumntype{L}[1]{>{\raggedright\let\newline\\\arraybackslash\hspace{0pt}}m{#1}}
\newcolumntype{C}[1]{>{\centering\let\newline\\\arraybackslash\hspace{0pt}}m{#1}}
\newcolumntype{R}[1]{>{\raggedleft\let\newline\\\arraybackslash\hspace{0pt}}m{#1}}
\usepackage{booktabs}
\def\toprule{\noalign{\smallskip\hrule height 1.2pt\smallskip}}
\def\midrule{\noalign{\smallskip\hrule\smallskip}}
\let\bottomrule=\toprule
\usepackage{makecell}

\usepackage[inline,shortlabels]{enumitem}

\usepackage{csquotes}

\usepackage{hyperref}

\usepackage[page,header]{appendix}

\usepackage{fancyhdr}
\begin{document}

% remove header and footer, as requested by camera-ready version
\fancyhf{}
\fancyhead{}
\fancyfoot{}

\title{Improving Contrastive Learning of Sentence Embeddings with Case-Augmented Positives and Retrieved Negatives}

\author{Wei Wang}
\email{zhuazhua.ww@alibaba-inc.com}
\orcid{0000-0003-0469-4938}
\affiliation{%
  \institution{Alibaba Group}
  \streetaddress{969 West Wen Yi Road}
  \city{Hangzhou}
  \state{Zhejiang}
  \country{China}
  \postcode{311121}
}

\author{Liangzhu Ge}
\email{liangzhu.glz@alibaba-inc.com}
\affiliation{%
  \institution{Alibaba Group}
  \streetaddress{969 West Wen Yi Road}
  \city{Hangzhou}
  \state{Zhejiang}
  \country{China}
  \postcode{311121}
}

\author{Jingqiao Zhang}
\email{jingqiao.zhang@alibaba-inc.com}
\affiliation{%
  \institution{Alibaba Group}
  \streetaddress{969 West Wen Yi Road}
  \city{Hangzhou}
  \state{Zhejiang}
  \country{China}
  \postcode{311121}
}

\author{Cheng Yang}
\email{charis.yangc@alibaba-inc.com}
\affiliation{%
  \institution{Alibaba Group}
  \streetaddress{969 West Wen Yi Road}
  \city{Hangzhou}
  \state{Zhejiang}
  \country{China}
  \postcode{311121}
}

\renewcommand{\shortauthors}{Wang and Ge, et al.}

\begin{abstract}
% TODO Need to deepen the idea: what kind of problem does switch case solve?
  Following SimCSE, contrastive learning based methods have achieved the state-of-the-art (SOTA) performance in learning sentence embeddings. However, the unsupervised contrastive learning methods still lag far behind the supervised counterparts. We attribute this to the quality of positive and negative samples, and aim to improve both. Specifically, for positive samples, we propose switch-case augmentation to flip the case of the first letter of randomly selected words in a sentence. This is to counteract the intrinsic bias of pre-trained token embeddings to frequency, word cases and subwords. For negative samples, we sample hard negatives from the whole dataset based on a pre-trained language model. Combining the above two methods with SimCSE, our proposed Contrastive learning with Augmented and Retrieved Data for Sentence embedding (CARDS) method significantly surpasses the current SOTA on STS benchmarks in the unsupervised setting. %Furthermore, we show on GLUE benchmark that switch-case augmentation is a general data augmentation technique for natural language understanding.
\end{abstract}




\maketitle

\section{Introduction}
It is a standard paradigm in natural language understanding to pre-train large-scale models, and adapt them to various downstream tasks \cite{survey,survey-prompt}. Recently, a contrastive learning framework called SimCSE \cite{simcse} is proposed to finetune the pre-trained BERT \cite{bert} and RoBERTa \cite{roberta} to learn sentence embeddings, with performance significantly surpassing previous results \cite{bert-whitening}. In the unsupervised setting, SimCSE encodes the same sentence twice with independent dropout noises to produce a positive pair of sentence embeddings. Meanwhile, it treats other samples from the same batch as negatives. SimCSE is further enhanced in follow-up studies by carefully designed data augmentation methods \cite{esimcse,vascl}, momentum contrast \cite{esimcse}, and prompt tuning \cite{prompt-bert}. Despite its success, there is still a significant gap between the performance of SimCSE trained using unlabelled English Wikipedia dataset and that using labelled SNLI+MNLI dataset \cite{snli,mnli}, where the latter has manually designed entailed and contradictory sentence pairs treated as positive and negative samples respectively (see Tab. \ref{Tab:cards_examples}). We attribute the gap to the higher quality of positive and negative samples in SNLI+MNLI than those produced by dropout and found in the same batch. %Indeed, if we remove the entailed and contradictory sentences in SNLI+MNLI dataset, the performance of SimCSE on semantic textual similarity (STS) tasks drops to \(73.8\), below that of Wikipedia (\(78.9\)) (see Tab. \ref{Tab:sts_add_pos_neg}). 
Thus, to narrow down the gap, we attempt to improve the quality of both positive and negative samples in the unsupervised setting.

Data augmentation is arguably the most straightforward approach to improve sample quality. However, it is known to be challenging for language data, possibly due to their discrete nature \cite{survey-augmentation}. 
%Existing data augmentation methods may be divided into three categories: rule-based approaches (e.g., dropout \cite{rdrop}, word insertion, deletion, swap \cite{eda}, crop \cite{cocolm}), MixUp interpolation \cite{mixup,seqmix}, and model-based techniques (e.g., back-translation \cite{back-translation,coda}, adversarial training \cite{freelb,advaug,alum}). In the context of sentence embedding learning, 
Existing rule-based methods, such as word insertion, deletion, swap \cite{eda} and crop \cite{cocolm}, may alter the semantic meaning of sentences, e.g., even deleting one word hurts the performance in the SimCSE framework \cite{simcse}.
Back-translation \cite{back-translation,coda} or other syntactic\slash semantic transformations \cite{augmentation-lit,augmentation-syntactic} can generate plausible samples, but may involve other models like GPT2 \cite{gpt2}. Adversarial training \cite{freelb,advaug,alum} tends to incur a large computational cost \cite{yopo}. 

%or other syntactic\slash semantic transformations

\begin{figure}[tb]
	\centering
	\begin{subfigure}[t]{0.21\textwidth}
		\centering
		\includegraphics[width=0.8\textwidth]{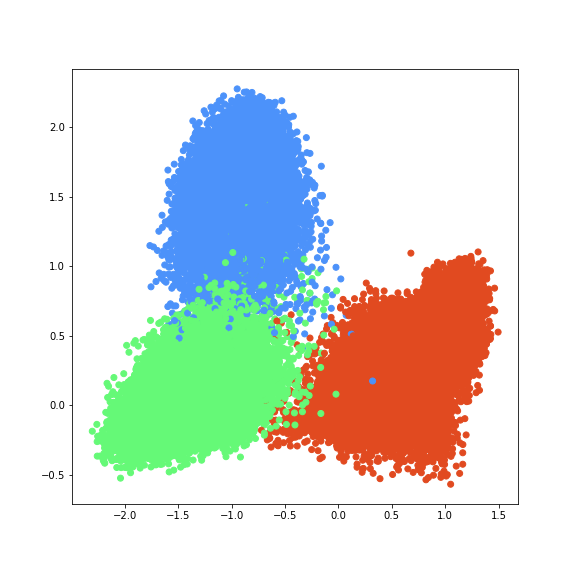}
		\caption{Case and sub-token biases.\label{fig:sub-token_case_bias}}
	\end{subfigure}
	~
	\begin{subfigure}[t]{0.21\textwidth}
		\centering
		\includegraphics[width=0.8\textwidth]{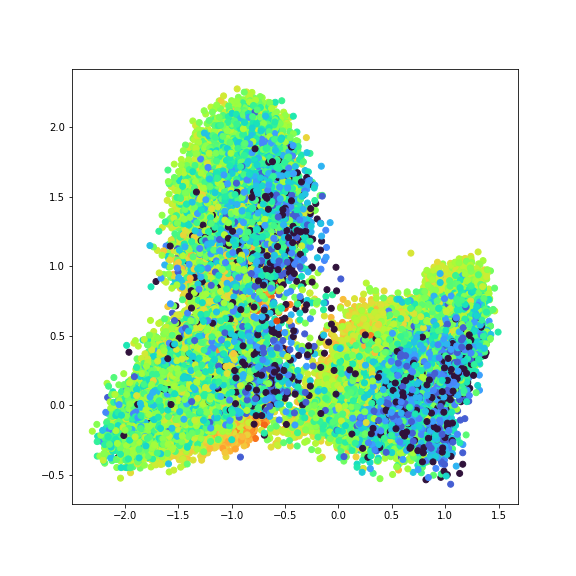}
		\caption{Frequency bias.\label{fig:freq_bias}}
	\end{subfigure}
	\caption{PCA visualization of \texorpdfstring{RoBERTa\textsubscript{large}}{RoBERTa-large} token embeddings with various biases. For (a) case and sub-token biases, green represents lower-case beginning tokens, blue upper-case beginning tokens, and red sub-tokens; for (b) frequency bias, the darker the color, the higher the frequency.}
	\label{fig:embd_biases}
\end{figure}

Is it possible to obtain positive samples with minimal semantic alternation and computational overhead? A promising answer, proposed by ESimCSE \cite{esimcse}, is to augment a sentence by randomly repeating some words\slash sub-words in it. 
%This alleviates the issue that the model may use sentence length to distinguish the negative pairs from the positives. 
However, token repetition may introduce unlikely samples. In natural language generation studies, token repetition is often considered as a degenerate property of Transformer-based models trained with the maximum likelihood objective \cite{nucleus-sampling,unlikelihood}. On the other hand, token embeddings in the pre-trained BERT and RoBERTa models are biased towards token frequency, word case and subwords \cite{bert-flow,prompt-bert,freq_bias_context_embds} (see Fig.\ref{fig:embd_biases}). The performance of using the average of token embeddings as the sentence embedding can be significantly improved by simply removing top frequent tokens, subwords and uppercase tokens \cite{prompt-bert}. Inspired by this, we propose switch-case augmentation: by flipping the case of the first letter of randomly selected words in a sentence, we change the frequency of tokens used, and in many cases, the tokenization of words and thus the length of sentence ids (see Tab. \ref{Tab:switch_case_examples}) - all is done without affecting much the meaning of the sentences from the human perspective.

As for negative samples, the in-batch local negatives may not provide enough informative information due to the diversity of natural sentences in a large corpus \cite{retrieval-approximate-nn,cl_temperature,cl_sample_hard_negatives}. %This is also indicated by fast decreasing loss curve of SimCSE during initial training (see Fig. \ref{fig:curve_log_loss}). 
ESimCSE expands the negative sample set with sentence embeddings from immediate preceding mini-batches using momentum contrast encoder \cite{moco}. We argue this may still not be effective, as the majority of sentences in Wikipedia corpus are trivially unrelated. Inspired by various text retrieval studies \cite{retrieval-approximate-nn,condenser,cocondenser,retrieval_dynamic_hard_negatives,retrieval-self-train,retrieval-retro}, we propose to search in the large corpus the most similar samples for each query sentence as its hard negatives. 

The above two optimizations, switch-case augmentation and negative retrieval, comprise our unsupervised learning method coined \textbf{C}ontrastive learning with \textbf{A}ugmented and \textbf{R}etrieved \textbf{D}ata for \textbf{S}entence embedding (CARDS). In the following sections, we conduct a comprehensive evaluation of CARDS on seven STS tasks for sentence embedding learning. %Furthermore, we test switch-case on GLUE benchmark datasets to show it is a general data augmentation approach for natural language understanding. 

\begin{table}[t]
	\caption{Effects of switch-case on RoBERTa tokenization.}
	\centering
	\label{Tab:switch_case_examples}
	\begin{threeparttable}
		\begin{tabular}{c c c}
			\toprule
			Type & Tokenization\tnote{1} & Percentage\tnote{2} \\
			\midrule
			\multirow{2}{*}{Substitution} & natural-istic \textrightarrow Natural-istic & 69.9 \\
			& Chart-ing \textrightarrow chart-ing & 15.0 \\
			\multirow{2}{*}{Division} & interpret \textrightarrow Inter-pret & 6.2 \\
			& Neigh-bor \textrightarrow ne-igh-bor & 3.9 \\
			\multirow{2}{*}{Fusion} & recomm-ended \textrightarrow Recommended & 1.8 \\
			& Ser-ious \textrightarrow serious & 1.3 \\
			\multirow{2}{*}{Regrouping} & urg-ency \textrightarrow Ur-gency & 1.2 \\
			& O-ng-oing \textrightarrow ongo-ing & 1.0 \\
			\bottomrule
		\end{tabular}
		\begin{tablenotes}
		    \item[1] The tokens of a word are connected by hyphens.
			\item[2] The percentage of occurrence is calculated on WiKi-1m, a subset of Wikipedia corpus used in SimCSE \cite{simcse}.
		\end{tablenotes}
	\end{threeparttable}
\end{table}

\section{Methods}
We first briefly present contrastive learning and SimCSE, then explain our proposed CARDS method in details.

\subsection{Contrastive Learning and SimCSE}

Contrastive learning is a general self-supervised learning framework. It works by maximizing the agreement between differently augmented views of the same data example, while separating the views of different examples \cite{simclr}. In the context of sentence embedding, each data example is a sentence \(x_i\). We denote \(f_\theta\) as the language encoder, and \(\bm{h}_i=f_\theta(x_i,\mathcal{T})\) as the embedding of \(x_i\) with augmentation \(\mathcal{T}\). SimCSE uses the following training objective:
\begin{equation}\label{eq:nce}
    \mathcal{l}_i = -\log{\frac{\exp(\text{cos}(\bm{h}_i,\bm{h}_i')/\tau)}{\sum_j^N\exp(\text{cos}(\bm{h}_i,\bm{h}_j')/\tau)}}
\end{equation}
where in unsupervised setting, the positive sample pair \((\bm{h}_i,\bm{h}_i')\) is obtained by passing \(x_i\) to the encoder twice with different dropout masks as data augmentation; for \(\bm{h}_i\), the \(N-1\) negative samples \(\bm{h}_{j=1:N,j\ne i}\) are simply from the same batch; the embedding similarity is measured using cosine operation and \(\tau\) is the temperature. %In the supervised setting, each sentence \(x_i\) has an entailed (or semantically similar) sentence \(x_i^+\) and a contradictory sentence \(x_i^-\) as the positive and additional negative samples respectively (see Tab. \ref{Tab:cards_examples} for examples). Acquired from the SNLI+MNLI dataset, these sentences are deliberately designed and manually annotated to make the task challenging for language models. %Thus, to improve the performance of SimCSE in the unsupervised setting, we devise methods to automatically enhance the quality of positive and negative samples, introduced below.

\subsection{Switch-case augmentation}

Previous studies find that the token embeddings of pre-trained BERT and RoBERTa are biased towards token frequency, word case and subword tokenization \cite{bert-flow,prompt-bert}. In Fig. \ref{fig:embd_biases}, we visualize the token embeddings of the pre-trained \texorpdfstring{RoBERTa\textsubscript{large}}{RoBERTa-large} model, where we call the first token of a word the beginning token, and the rest subordinate tokens or sub-tokens. For example, `ne-igh-bor' is tokenized into one beginning token `ne' and two sub-tokens `igh' and `bor'. Fig. \ref{fig:sub-token_case_bias} shows an almost clear separation of embeddings for lower-case (green), upper-case (blue) and sub-tokens (red); Fig. \ref{fig:freq_bias} shows gradual change from rare token (light green) regions to frequent token (dark blue) regions for each cluster of tokens. Moreover, these biases exist in contextualized token embeddings (the hidden states of last few layers) of pre-trained BERT, and additional data fails to mitigate the bias towards word frequency \cite{freq_bias_context_embds}. 

\begin{table}[t]
	\caption{Examples of positive and negative sentences.}
	\centering
	\label{Tab:cards_examples}
	\begin{threeparttable}
		\begin{tabular}{l L{6cm}}
			\toprule
			Type & Sentence (NO. tokens) \\
			\midrule
			Original & The story of the first book continues. (8) \\
			Case-switched & The story of the first book Continues. (9) \\
			Entailed\tnote{1} & The story from the first book goes on. (9) \\
			\midrule
			Retrieved & The story begins as a typical love story. (9) \\
			& The book tells the story of a curious rabbit. (10) \\
			Random & This is held as a temporary result. (8) \\
			%& In total she won four gold medals and three bronze medals. (12) \\
			Contradictory\tnote{1} & The story of the first book ends. (8) \\
			\bottomrule
		\end{tabular}
		\begin{tablenotes}
			\item[1] The entailed and contradictory sentences shown here are manually designed to mimic those in the SNLI+MNLI dataset.
		\end{tablenotes}
	\end{threeparttable}
\end{table}

We propose switch-case augmentation to alleviate the biases. The idea is simple: we randomly select words in a sentence with a fixed probability \(p_{sc}\) and flip the case of the first letter of these words. There are four possible consequences (see Tab. \ref{Tab:switch_case_examples}):
\begin{enumerate}[leftmargin=*]
	\item Substitution. Only the beginning token is replaced with a case-switched token; the sub-tokens are not affected. 
	\item Division. The case-switched subword is tokenized into two or more tokens, thus the total number of tokens \(N\) increases.
	\item Fusion. The case-switched subword is combined with other tokens into one, thus \(N\) decreases.
	\item Regrouping. The case-switched subword is regrouped with other tokens, and \(N\) may increase, decrease or remain the same.
\end{enumerate}
Tab. \ref{Tab:switch_case_examples} also lists the occurring proportion of each consequence when applying switch-case to sentences in Wiki-1m dataset, a subset of Wikipedia corpus used in SimCSE study \cite{simcse}. In about \(85\%\) of the cases, the beginning token is replaced with another token of probably different frequency. In \(14\%\) of the cases, the total number of tokens varies. ESimCSE \cite{esimcse} randomly repeats some tokens to avoid the trivial solution of using sentence length to distinguish the negative pairs from the positives. We achieve this with a negligible influence on the sentence semantics (see Tab. \ref{Tab:cards_examples} for an example).%: the case-switched sentence has the same meaning as the original sentence, but different number of tokens as `Continues' is divided into `Contin' and `ues'). 

\begin{table*}[t]
	\caption{Test performance of unsupervised sentence embedding on STS tasks.}
	\centering
	\label{Tab:sts_test}
	\begin{threeparttable}
		\begin{tabular}{l l c c c c c c c l}
			\toprule
			Base & Model & STS12 & STS13 & STS14 & STS15 & STS16 & STS-B & SICK-R & Avg \\
			\midrule
			\multirow{8}{*}{\texorpdfstring{RoBERTa\textsubscript{base}}{RoBERTa-base}} & SimCSE\tnote{1} & 70.16 & 81.77 & 73.24 & 81.36 & 80.65 & 80.22 & 68.56 & 76.57 (-0.43) \\
			& ESimCSE \cite{esimcse} & 69.90 & 82.50 & 74.68 & 83.19 & 80.30 & 80.99 & 70.54 & 77.44 (+0.44) \\
			& VaSCL \cite{vascl} & 69.08 & 81.95 & 74.64 & 82.64 & 80.57 & 80.23 & 71.23 & 77.19 (+0.19) \\
			& Prompt-RoBERTa \cite{prompt-bert}\tnote{2} & \textbf{73.94} & \textbf{84.74} & \textbf{77.28} & \textbf{84.99} & 81.74 & 81.88 & 69.50 & \textbf{79.15} (+2.15) \\
			\cmidrule{2-10}
			& SimCSE (ours)\tnote{3} & 68.86 & 82.21 & 73.39 & 81.43 & 81.80 & 81.00 & 70.34 & 77.00 \\
			& \(\;\;+\)switch-case & 70.62 & 82.11 & 75.00 & 81.88 & 81.40 & 81.92 & \textbf{71.52} & 77.78 (+0.78) \\
			& \(\;\;+\)retrieval & 70.33 & 82.78 & 74.54 & 82.64 & 81.79 & 81.16 & 69.26 & 77.50 (+0.50) \\
			& CARDS\tnote{4} & 72.49 & 84.09 & 76.19 & 82.98 & \textbf{82.11} & \textbf{82.25} & 70.65 & 78.68 (+1.68) \\
			\midrule
			\multirow{7}{*}{\texorpdfstring{RoBERTa\textsubscript{large}}{RoBERTa-large}} & SimCSE\tnote{1} & 72.86 & 83.99 & 75.62 & 84.77 & 81.80 & 81.98 & 71.26 & 78.90 \\
			& ESimCSE \cite{esimcse} & 73.20 & 84.93 & 76.88 & 84.86 & 81.21 & 82.79 & 72.27 & 79.45 (+0.55)\\
			& VaSCL \cite{vascl} & 74.34 & 83.35 & 76.79 & 84.37 & 81.46 & 82.86 & \textbf{73.23} & 79.48 (+0.58) \\
			\cmidrule{2-10}
			& SimCSE (ours)\tnote{3} & 71.51 & 83.21 & 76.06 & 84.12 & 82.16 & 82.31 & 71.80 & 78.74 (-0.16) \\
			& \(\;\;+\)switch-case & 73.30 & 84.58 & 77.16 & 84.89 & 81.78 & 82.90 & 71.88 & 79.50 (+0.60) \\
			& \(\;\;+\)retrieval & 73.41 & 85.27 & 77.76 & 85.44 & 82.19 & 82.54 & 70.87 & 79.64 (+0.74) \\
			& CARDS\tnote{4} & \textbf{74.63} & \textbf{86.27} & \textbf{79.25} & \textbf{85.93} & \textbf{83.17} & \textbf{83.86} & 72.77 & \textbf{80.84} (+1.94)\\
			\bottomrule
		\end{tabular}
		\begin{tablenotes}
			\item[1] Results obtained from GitHub checkpoints: https://github.com/princeton-nlp/SimCSE. \item[2] \texorpdfstring{Prompt-RoBERTa\textsubscript{large}}{Prompt-RoBERTa-large} results are not officially released. Upon paper submission, we were unable to achieve promising results using its code and manually designed prompts. \item[3] Our reproduced SimCSE results are different from that of SimCSE GitHub. We use the higher ones as baseline to calculate the relative improvements of each method in brackets. \item[4] CARDS = SimCSE \(+\) switch-case \(+\) retrieval.
		\end{tablenotes}
	\end{threeparttable}
\end{table*}

\subsection{Hard negative retrieval}

%In this section, we consider improving the quality of negative samples by searching for hard negatives that are difficult to distinguish from the query samples \cite{cl_temperature,cl_sample_hard_negatives}. We argue that 
The in-batch negatives, sampled uniformly from the training corpus, may not be hard enough due to the diversity of natural sentences \cite{cl_temperature,cl_sample_hard_negatives}. Inspired by text retrieval studies \cite{retrieval_dynamic_hard_negatives,retrieval-approximate-nn}, we propose to retrieve top \(k\) hard negatives in the training corpus for each sample \(x_i\) in the current batch, uniformly sample \(s=1\) from them to get \(x_i^-\), and use it in the training objective: 
\begin{equation}\label{eq:nce_hard_neg}
    \mathcal{l}_i = -\log{\frac{\exp(\text{cos}(\bm{h}_i,\bm{h}_i')/\tau)}{\sum_j^N[\exp(\text{cos}(\bm{h}_i,\bm{h}_j')/\tau)+\exp(\text{cos}(\bm{h}_i,\bm{h}_j^-)/\tau)]}}
\end{equation}
Note \(x_i^-\) will be treated as a random negative for other samples in the batch. To do retrieval, we first build the index, or the representation of each sentence in the training corpus, by passing each one to the same pre-trained language model to be finetuned in contrastive learning. Then the hardness is measured as cosine similarity between each pair of representations. See Tab. \ref{Tab:cards_examples} for examples of retrieved and random negatives.

\section{Experiments}

\subsection{Evaluation Setup}

For sentence embedding learning, we follow the same process as SimCSE-related studies \cite{simcse,esimcse,vascl,prompt-bert} to evaluate our proposed CARDS framework, and compare it against SimCSE \cite{simcse}, ESimCSE \cite{esimcse}, VaSCL \cite{vascl} and PromptBERT \cite{prompt-bert}. Later in Appendix \ref{appendix:glue}, we also evaluate switch-case on \texorpdfstring{DeBERTa\textsubscript{1.5B}}{DeBERTa-1.5B} and the GLUE benchmark to show its generality as a data augmentation approach.

\paragraph{Datasets} We use Wiki-1m from SimCSE as our training set, which consists of 1 million sentences randomly drawn from the English Wikipedia corpus. We use STS12-STS16 \cite{sts12,sts13,sts14,sts15,sts16} and STS-B \cite{stsb} from the SentEval toolkit \cite{senteval} as our evaluation sets. For each STS task, the standard evaluation pipeline from SentEval is used.
%Wiki-1m consists of 1 million sentences randomly drawn from the English Wikipedia corpus. %SentEval is an evaluation tookit comprising a wide variety of language tasks including sentiment analysis, entailment prediction, STS, paraphrase detection, caption-image retrieval, etc. Among them, STS tasks are selected due to its direct comparison of sentence embeddings.  
%SentEval provides standard evaluation pipelines for its tasks.
%: for STS12-STS16 \cite{sts12,sts13,sts14,sts15,sts16} and STS-B \cite{stsb}, the ranks of cosine similarities of sentence pairs in the test set are directly compared with ground truth labels using the Spearman correlation; for SICK-R \cite{sickr}, an extra linear classifier is inserted and trained to predict the relatedness score before assessing the Spearman correlation. 

%\paragraph{Baselines} The proposed CARDS is built on and thus compared with SimCSE \cite{simcse}. In addition, we compare CARDS with ESimCSE \cite{esimcse} and VaSCL \cite{vascl} as they also propose to enhance the quality of positive and negative samples. At last, we consider PromptBERT \cite{prompt-bert} as an orthogonal method that finetunes BERT and RoBERTa with prompts, and may be combined with ours to further improve contrastive learning of sentence embeddings.

\paragraph{Implementation details} We finetune the pre-trained checkpoints of \texorpdfstring{RoBERTa\textsubscript{base}}{RoBERTa-base} and \texorpdfstring{RoBERTa\textsubscript{large}}{RoBERTa-large} on Wiki-1m for one epoch. The maximum number of tokens in a sentence is confined to 32. Meanwhile, we evaluate the models on STS-B development dataset every 125 steps to select the best intermediate model for test. %At last, the models are assessed using the standard pipeline in SentEval. 
For switch-case, we select \(p_{sc}\) in \(\{0.05,0.1,0.15\}\) and fix dropout rate at \(0.1\). For hard negative retrieval, we adopt Faiss library\footnote{\href{https://github.com/facebookresearch/faiss}{https://github.com/facebookresearch/faiss}} \cite{faiss} to efficiently build the index and do similarity search. The index or sentence representations for retrieval are calculated using the pre-trained checkpoints before any finetuning and shared across all experiments. For each sample, the number of hard negatives \(k\) to retrieve (excluding itself) is selected from \(\{8,64\}\). We also deduplicate the Wiki-1m dataset and remove sentences with less than three words. This does not affect much the final performance but significantly stabilize the training in the late stages. Our code\footnote{\href{https://github.com/alibaba/SimCSE-with-CARDS}{https://github.com/alibaba/SimCSE-with-CARDS}.} is implemented based on the HuggingFace Transformers library\footnote{\href{https://github.com/huggingface/transformers}{https://github.com/huggingface/transformers}} \cite{huggingface}.

\subsection{Main Results}

%\subsubsection{Sentence embedding learning}
As shown in Tab. \ref{Tab:sts_test}, the proposed CARDS significantly improve over SimCSE, ESimCSE and VaSCL on almost all STS tasks and the average STS scores for both RoBERTa-base and RoBERTa-large models. Moreover, adding either switch-case or retrieval to SimCSE could achieve comparable average STS scores to ESimCSE and VaSCL. %In brackets, we also list the difference between each method and baselines. CARDS outperforms SimCSE by 1.68 on RoBERT-base and 1.94 on -large. 
It is interesting to note that combining switch-case and negative retrieval in CARDS achieves even a bigger improvement than simply summing up their respective improvements together, showing that the two methods may strengthen the effects of each other. Although on \texorpdfstring{RoBERTa\textsubscript{base}}{RoBERTa-base}, CARDS does not perform as good as Prompt-RoBERTa, we argue that Prompt-RoBERTa is an orthogonal method in that it finetunes RoBERTa with prompts, and may be combined with CARDS. 

%The proposed method works towards closing the large gaps between unsupervised and supervised SimCSE (the gaps being 5.52 and 4.86 respectively). We show the loss curves (in logarithmic scale) and the evaluation performances on STS-B dataset of SimCSE and CARDS during training in Fig. \ref{fig:curve_log_loss_stsb}. It indicates that by augmenting the positive and retrieving the negatives, the contrastive learning task indeed becomes harder, and the model performance increases much faster especially during the early training stage. We also list the incremental improvements of adding higher-quality positives and negatives using our methods and the supervised SNLI+MNLI dataset in Tab \ref{Tab:sts_add_pos_neg}. The manually designed entailed and contradictory sentences, when used as positives and negatives respectively, add 6.93 and 4.21 each and 9.66 in total to the average STS score of baseline model. Note this can be partially attributed to the fact that STS data are more similar to SNLI and MNLI than Wiki-1m. However, this also indicates that future work should further investigate the quality of positive and negative samples. 

\subsection{Ablation studies}
\label{sec:ablation}

\begin{figure}[t]
	\centering
	\begin{subfigure}[t]{0.5\linewidth}
		\centering
    	\begin{tikzpicture}
    	\begin{axis}[%
    	no markers,
    	width=0.9\linewidth,
    	axis y line*=left,
    	xmin=0.05, xmax=0.20, ymin=76.5, ymax=78.0, tick label style={font=\tiny},
    	xticklabel style={
        /pgf/number format/fixed,
        /pgf/number format/precision=2
        },
        scaled x ticks=false,
        yticklabel style={color=turbo1},
    	xlabel=\(p_{cs}\), ylabel=STS Avg., 
    	label style={font=\footnotesize},
    	ylabel style={color=turbo1},
    	legend columns=-1,
    	legend style={at={(0,1)},anchor=south west, font=\small, draw=none, row sep=-4.0pt, inner sep=0pt}
    	]
    	\addplot[color=turbo1,line width=2pt]table[x=x, y=y0] {figures/hp_psc_roberta_base.dat};
    	\addplot[dashed,dash phase=1pt,color=turbo1,line width=1pt] coordinates {(0.05,77) (0.2,77)};
    	%\legend{SimCSE (unsupervised),CARDS,SimCSE (supervised)};
    	\end{axis}
    	
    	\begin{axis}[%
    	no markers,
    	width=0.9\linewidth,
    	axis x line=none,
    	axis y line*=right,
    	xmin=0.05, xmax=0.20, ymin=545, ymax=570, tick label style={font=\tiny},
    	yticklabel style={color=turbo2},
    	ylabel=Walltime (sec), 
    	label style={font=\footnotesize},
    	ylabel style={color=turbo2},
    	legend columns=-1,
    	legend style={at={(0,1)},anchor=south west, font=\small, draw=none, row sep=-4.0pt, inner sep=0pt}
    	]
    	\addplot[color=turbo2,line width=2pt]table[x=x, y=y1] {figures/hp_psc_roberta_base.dat};
    	\addplot[dashed,dash phase=1pt,color=turbo2,line width=1pt] coordinates {(0.05,548) (0.2,548)};
    	%\legend{SimCSE (unsupervised),CARDS,SimCSE (supervised)};
    	\end{axis}
    	\end{tikzpicture}
    	\caption{Switch-case probability.\label{fig:hp_psc}}
	\end{subfigure}
	~
	\begin{subfigure}[t]{0.5\linewidth}
		\centering
    	\begin{tikzpicture}
    	\begin{axis}[%
    	no markers,
    	width=0.9\linewidth,
    	axis y line*=left,
    	xmin=3, xmax=10, ymin=76.5, ymax=78.0, tick label style={font=\tiny},
    	xticklabel style={
        /pgf/number format/fixed,
        /pgf/number format/precision=2
        },
        scaled x ticks=false,
        yticklabel style={color=turbo1},
    	xlabel=\(\log_2{k}\), ylabel=STS Avg., 
    	label style={font=\footnotesize},
    	ylabel style={color=turbo1},
    	legend columns=-1,
    	legend style={at={(0,1)},anchor=south west, font=\small, draw=none, row sep=-4.0pt, inner sep=0pt}
    	]
    	\addplot[color=turbo1,line width=2pt]table[x=x, y=y0] {figures/hp_k_roberta_base.dat};
    	\addplot[dashed,dash phase=1pt,color=turbo1,line width=1pt] coordinates {(3,77) (10,77)};
    	%\legend{SimCSE (unsupervised),CARDS,SimCSE (supervised)};
    	\end{axis}
    	
    	\begin{axis}[%
    	no markers,
    	width=0.9\linewidth,
    	axis x line=none,
    	axis y line*=right,
    	xmin=3, xmax=10, ymin=520, ymax=800, tick label style={font=\tiny},
    	yticklabel style={color=turbo2},
    	ylabel=Walltime (sec), 
    	label style={font=\footnotesize},
    	ylabel style={color=turbo2},
    	legend columns=-1,
    	legend style={at={(0,1)},anchor=south west, font=\small, draw=none, row sep=-4.0pt, inner sep=0pt}
    	]
    	\addplot[color=turbo2,line width=2pt]table[x=x, y=y1] {figures/hp_k_roberta_base.dat};
    	\addplot[dashed,dash phase=1pt,color=turbo2,line width=1pt] coordinates {(3,548) (10,548)};
    	%\legend{SimCSE (unsupervised),CARDS,SimCSE (supervised)};
    	\end{axis}
    	\end{tikzpicture}
    	\caption{Number of retrieved negatives.\label{fig:hp_k}}
	\end{subfigure}
	\caption{Effects of hyper-parameters on average STS test scores and training walltimes\protect\footnotemark of \texorpdfstring{RoBERTa\textsubscript{base}}{RoBERTa-base}. The baseline score and walltime are shown in dashed lines. \label{fig:effects_hp}}
\end{figure}
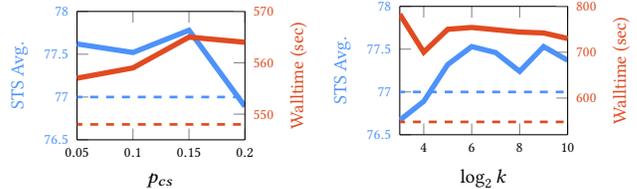

\begin{table}[t]
	\caption{Average STS test scores of \texorpdfstring{SimCSE-RoBERTa\textsubscript{large}}{SimCSE-RoBERTa-large} with upper-case ignored during training/evaluation.}
	\centering
	\label{Tab:remove_upper}
	\begin{threeparttable}
		\begin{tabular}{c c c c c}
			\toprule
			\multirow{2}{*}{Baseline} & \multicolumn{3}{c}{Ignoring upper-case during\tnote{1}} & \multirow{2}{*}{Switch-case} \\
			& Training & Evaluation & Train. \& Eval. & \\
			\midrule
			78.74 & 78.61 & 78.67 & 78.70 & 79.50 \\
			\bottomrule
		\end{tabular}
		\begin{tablenotes}
			\item[1] Among all STS tasks, only SICK-R involves upper-case words. Thus, to enlarge the impact of upper-case words, we switch the first letter of each sentence to upper case for all STS tasks.
		\end{tablenotes}
	\end{threeparttable}
\end{table}

\footnotetext{Note the walltimes also depend on hardware and code implementation. They are shown here for a rough illustration of the method complexity and should not be used to verify user implementation.}

In this section, we investigate how the performance of CARDS on STS tasks is influenced by design and hyper-parameters of the proposed switch-case and negative retrieval.

For switch-case, Fig. \ref{fig:hp_psc} shows that it adds a negligible overhead to the training walltime and increasing the switch-case probability \(p_{sc}\) above 0.15 will deteriorate the average STS score. In Tab. \ref{Tab:remove_upper}, we test whether ignoring the upper-case words (by switching them to lower case) during training and\slash or evaluation affects the performance. The results indicate that switch-case works not because it makes the model insensitive to word case. We further investigate whether the success of switch-case is mainly due to the substitution of tokens or the change of tokenization. For this, we consider two augmentation approaches: \begin{enumerate*}
  \item \emph{substitution} - do switch-case only for words of the substitution type (see Tab. \ref{Tab:switch_case_examples}) to remove the effects of varied sequence length; %this approach is denoted as substitution;
  \item \emph{re-tokenization} - for division and regrouping-type words, switch their case, keep the changed tokenization, then switch their case back and do tokenization again, so that their case is not changed but tokenization varies, e.g., `interpret' is forced to be tokenized as `inter' and `pret'.
\end{enumerate*}
The re-tokenization approach is similar to BPE-dropout \cite{bpe-dropout} which corrupts the segmentation process of BPE to provide several tokenizations, but BPE-dropout is applied to all words. Tab. \ref{Tab:variants_sc} shows that both substitution and re-tokenization improve over the baseline, while re-tokenization performs similar to the default switch-case, showing that switch-case works mainly due to the change of tokenization of certain long words\footnote{Interestingly, BPE-dropout generally hurts the performance across different hyper-parameter settings in our SimCSE framework, but ignoring BPE-dropout for the most frequent words hurts less.}. Given this, the performance drop when raising \(p_{sc}\) above 0.15 in Fig. \ref{fig:hp_psc} may be caused by the increasing difficulty of associating more tokens with their retokenized version. At last, we show in Appendx~\ref{appendix:glue} that switch-case is a general data augmentation approach applicable to \texorpdfstring{DeBERTa\textsubscript{1.5B}}{DeBERTa-1.5B} on GLUE tasks.

Regarding negative retrieval, Fig. \ref{fig:hp_k} shows that it increases roughly 40\% of the training time, partially because negative retrieval doubles the number of negative samples. Increasing the number of retrieved negatives do not vary the training time too much, possibly due to the efficient implementation of retrieval in Faiss, but decreasing it below 32 (\(=2^5\)) significantly deteriorates the performance. We further investigates the effects of the number of sampled negatives \(s\) from the retrieved ones and the type of negatives used in contrastive learning. For the latter one, we consider three types: \begin{enumerate*}
  \item the default used in CARDS, denoted as \(\mathbb{R}_{\text{uniform}}\), samples \(s\) negatives uniformly from \(k\) retrieved ones;
  \item \(\mathbb{R}_{\text{top}}\) selects only the top \(s\), or \(s\) hardest negatives from the retrieved ones; and 
  \item \(\mathbb{D}_{\text{uniform}}\) samples \(s\) negatives uniformly from the whole training set. 
\end{enumerate*}
Fig. \ref{fig:variants_retrieval} shows that both increasing \(s\) beyond 1 in \(\mathbb{R}_{\text{uniform}}\) and selecting the \(s\) hardest negatives in \(\mathbb{R}_{\text{top}}\) significantly deteriorate the performance, possibly due to the increasingly adverse impact of false negatives. On the other hand, sampling uniformly from the whole dataset only helps when \(s\) is large, which, however, incurs a considerable computational cost due to more negative samples. Currently, we sample uniformly from \(k\) retrieved ones to balance the difficulty of hard negatives and the impact of false negatives. We suspect filtering out false negatives  \cite{false_negative_detection} may further help, and leave this for future work. 

\begin{table}[t]
    \begin{minipage}[h]{.45\linewidth}
    \captionof{table}{Average STS test scores of \texorpdfstring{RoBERTa\textsubscript{base}}{RoBERTa-base} with three switch-case variants (default, substitution and re-tokenization) and BPE-dropout.\label{Tab:variants_sc}}
    \centering
    	\begin{threeparttable}
    		\begin{tabular}{l c}
    			\toprule
    			Variants & STS Avg. \\
    			\midrule
    			Baseline & 77.00 \\
    			\midrule
    			Default & 77.78 \\
    			Substit. & 77.43 \\
                Re-token. & 77.70 \\
                \midrule
                BPE-drop. & 76.74 \\
    			\bottomrule
    		\end{tabular}
    		\iffalse
    		\begin{tablenotes}
    			\item[1] 
    		\end{tablenotes}
    		\fi
    	\end{threeparttable}
    \end{minipage}
    ~
    \begin{minipage}[h]{.5\linewidth}
    \centering
    	\begin{tikzpicture}
    	\begin{axis}[%
    	no markers,
    	width=1.1\linewidth,
    	height=0.8\linewidth,
    	xmin=1, xmax=5, ymin=75.5, ymax=77.6, tick label style={font=\tiny},
    	xlabel=Number of sampled negatives, ylabel=STS Avg., label style={font=\footnotesize},
    	%legend style={at={(1,0.55)},anchor=north east, font=\footnotesize, draw=none, row sep=-4.0pt, inner sep=0pt},
    	% legend cell align={left}
    	legend style={at={(-0.14,1.05)},anchor=south west, font=\small, draw=none, row sep=-4.0pt, inner sep=0pt},
    	legend columns=-1,
    	]
    	\addplot[color=turbo1,line width=2pt]table[x=x, y=rand] {figures/variants_retrieval.dat};
    	\addplot[color=turbo2,line width=2pt]table[x=x, y=default] {figures/variants_retrieval.dat};
    	\addplot[color=turbo4,line width=2pt]table[x=x, y=top] {figures/variants_retrieval.dat};
    	\legend{\(\mathbb{D}_{\text{uniform}}\),\(\mathbb{R}_{\text{uniform}}\),\(\mathbb{R}_{\text{top}}\)};
    	\addplot[dashed,dash phase=1pt,color=turbo1,line width=1pt] coordinates {(1,77) (5,77)};
    	\end{axis}
    	\end{tikzpicture}
    	
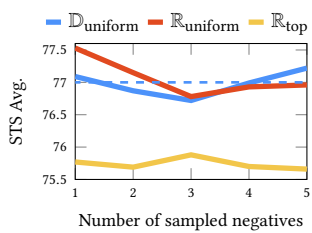
\captionof{figure}{Effects of number of sampled negatives on average STS test scores using \texorpdfstring{RoBERTa\textsubscript{base}}{RoBERTa-base} and negative retrieval variants. The baseline is shown in dashed line.\label{fig:variants_retrieval}}
    \end{minipage}
\end{table}

% \iffalse
\subsection{Negative results}
\label{sec:negative_results} % if add glue, remove this
Below we briefly describe two ideas that did not look promising in our initial experiments:
\begin{enumerate}[leftmargin=*]
  \item We applied switch-case to \texorpdfstring{BERT\textsubscript{large-cased}}{BERT-large-cased} but to no avail. We suspect that perhaps due to the relatively small vocabulary of BERT (29k), case-switched words tend to be tokenized to smaller tokens, making it harder to learn the token embeddings. It should also be noted that switch-case cannot be applied to uncased models such as \texorpdfstring{BERT\textsubscript{large-uncased}}{BERT-large-uncased}. However, switch-case can be applied as a general data augmentation approach to \texorpdfstring{DeBERTa\textsubscript{1.5B}}{DeBERTa-1.5B} on GLUE tasks, as shown in Appendx~\ref{appendix:glue}.
  \item Note that \cite{retrieval_dynamic_hard_negatives,retrieval-approximate-nn} propose to update the index during training to get ``dynamic" hard negatives. However, we tried various update strategies but none improved over the static one. We suspect that, with training going on, the false negatives become more frequently retrieved and deteriorate the training. Thus, we only build the index once before training and do the retrieval using fixed sentence embeddings, and leave ``dynamic" hard negatives retrieval for future study. 
  %\item We noticed a lot of wrong sentence breaks in the Wiki-1m corpus. Then we designed rules to correctly split the paragraphs in Wikipedia dataset into sentences and sampled 1m sentences, leading to a new dataset with much longer average sentence length than Wiki-1m. This new dataset, however, did not lead to a better performance on STS. 
\end{enumerate}
% \fi

\section{Conclusions}
In this study, we propose Contrastive learning with Augmented and Retrieved Data for Sentence embedding (CARDS), which introduces switch-case augmentation and hard negative retrieval to improve the positive and negative samples respectively in the SimCSE framework. We show that CARDS works towards closing the gap between unsupervised and supervised training of SimCSE. We believe that following our work, more data augmentation methods could be tailored to further close this gap.  

\bibliographystyle{ACM-Reference-Format}
\bibliography{references}


\begin{thebibliography}{54}

%%% the Web address.

\ifx \showCODEN    \undefined \def \showCODEN     #1{\unskip}     \fi
\ifx \showDOI      \undefined \def \showDOI       #1{#1}\fi
\ifx \showISBNx    \undefined \def \showISBNx     #1{\unskip}     \fi
\ifx \showISBNxiii \undefined \def \showISBNxiii  #1{\unskip}     \fi
\ifx \showISSN     \undefined \def \showISSN      #1{\unskip}     \fi
\ifx \showLCCN     \undefined \def \showLCCN      #1{\unskip}     \fi
\ifx \shownote     \undefined \def \shownote      #1{#1}          \fi
\ifx \showarticletitle \undefined \def \showarticletitle #1{#1}   \fi
\ifx \showURL      \undefined \def \showURL       {\relax}        \fi
% invisible to TeX
\providecommand\bibfield[2]{#2}
\providecommand\bibinfo[2]{#2}
\providecommand\natexlab[1]{#1}
\providecommand\showeprint[2][]{arXiv:#2}

\bibitem[Agirre et~al\mbox{.}(2015)]%
        {sts15}
\bibfield{author}{\bibinfo{person}{Eneko Agirre}, \bibinfo{person}{Carmen
  Banea}, \bibinfo{person}{Claire Cardie}, \bibinfo{person}{Daniel Cer},
  \bibinfo{person}{Mona Diab}, \bibinfo{person}{Aitor Gonzalez-Agirre},
  \bibinfo{person}{Weiwei Guo}, \bibinfo{person}{I{\~n}igo Lopez-Gazpio},
  \bibinfo{person}{Montse Maritxalar}, \bibinfo{person}{Rada Mihalcea},
  \bibinfo{person}{German Rigau}, \bibinfo{person}{Larraitz Uria}, {and}
  \bibinfo{person}{Janyce Wiebe}.} \bibinfo{year}{2015}\natexlab{}.
\newblock \showarticletitle{{S}em{E}val-2015 Task 2: Semantic Textual
  Similarity, {E}nglish, {S}panish and Pilot on Interpretability}. In
  \bibinfo{booktitle}{\emph{Proceedings of the 9th International Workshop on
  Semantic Evaluation ({S}em{E}val 2015)}}. \bibinfo{pages}{252--263}.
\newblock


\bibitem[Agirre et~al\mbox{.}(2014)]%
        {sts14}
\bibfield{author}{\bibinfo{person}{Eneko Agirre}, \bibinfo{person}{Carmen
  Banea}, \bibinfo{person}{Claire Cardie}, \bibinfo{person}{Daniel Cer},
  \bibinfo{person}{Mona Diab}, \bibinfo{person}{Aitor Gonzalez-Agirre},
  \bibinfo{person}{Weiwei Guo}, \bibinfo{person}{Rada Mihalcea},
  \bibinfo{person}{German Rigau}, {and} \bibinfo{person}{Janyce Wiebe}.}
  \bibinfo{year}{2014}\natexlab{}.
\newblock \showarticletitle{{S}em{E}val-2014 Task 10: Multilingual Semantic
  Textual Similarity}. In \bibinfo{booktitle}{\emph{Proceedings of the 8th
  International Workshop on Semantic Evaluation ({S}em{E}val 2014)}}.
  \bibinfo{pages}{81--91}.
\newblock


\bibitem[Agirre et~al\mbox{.}(2016)]%
        {sts16}
\bibfield{author}{\bibinfo{person}{Eneko Agirre}, \bibinfo{person}{Carmen
  Banea}, \bibinfo{person}{Daniel Cer}, \bibinfo{person}{Mona Diab},
  \bibinfo{person}{Aitor Gonzalez-Agirre}, \bibinfo{person}{Rada Mihalcea},
  \bibinfo{person}{German Rigau}, {and} \bibinfo{person}{Janyce Wiebe}.}
  \bibinfo{year}{2016}\natexlab{}.
\newblock \showarticletitle{{S}em{E}val-2016 Task 1: Semantic Textual
  Similarity, Monolingual and Cross-Lingual Evaluation}. In
  \bibinfo{booktitle}{\emph{Proceedings of the 10th International Workshop on
  Semantic Evaluation ({S}em{E}val-2016)}}. \bibinfo{pages}{497--511}.
\newblock


\bibitem[Agirre et~al\mbox{.}(2012)]%
        {sts12}
\bibfield{author}{\bibinfo{person}{Eneko Agirre}, \bibinfo{person}{Daniel Cer},
  \bibinfo{person}{Mona Diab}, {and} \bibinfo{person}{Aitor Gonzalez-Agirre}.}
  \bibinfo{year}{2012}\natexlab{}.
\newblock \showarticletitle{{S}em{E}val-2012 Task 6: A Pilot on Semantic
  Textual Similarity}. In \bibinfo{booktitle}{\emph{*{SEM} 2012: The First
  Joint Conference on Lexical and Computational Semantics {--} Volume 1:
  Proceedings of the main conference and the shared task, and Volume 2:
  Proceedings of the Sixth International Workshop on Semantic Evaluation
  ({S}em{E}val 2012)}}. \bibinfo{pages}{385--393}.
\newblock


\bibitem[Agirre et~al\mbox{.}(2013)]%
        {sts13}
\bibfield{author}{\bibinfo{person}{Eneko Agirre}, \bibinfo{person}{Daniel Cer},
  \bibinfo{person}{Mona Diab}, \bibinfo{person}{Aitor Gonzalez-Agirre}, {and}
  \bibinfo{person}{Weiwei Guo}.} \bibinfo{year}{2013}\natexlab{}.
\newblock \showarticletitle{*{SEM} 2013 shared task: Semantic Textual
  Similarity}. In \bibinfo{booktitle}{\emph{Second Joint Conference on Lexical
  and Computational Semantics (*{SEM}), Volume 1: Proceedings of the Main
  Conference and the Shared Task: Semantic Textual Similarity}}.
  \bibinfo{pages}{32--43}.
\newblock


\bibitem[Borgeaud et~al\mbox{.}(2021)]%
        {retrieval-retro}
\bibfield{author}{\bibinfo{person}{Sebastian Borgeaud}, \bibinfo{person}{Arthur
  Mensch}, \bibinfo{person}{Jordan Hoffmann}, \bibinfo{person}{Trevor Cai},
  \bibinfo{person}{Eliza Rutherford}, \bibinfo{person}{Katie Millican},
  \bibinfo{person}{George van~den Driessche}, \bibinfo{person}{Jean{-}Baptiste
  Lespiau}, \bibinfo{person}{Bogdan Damoc}, \bibinfo{person}{Aidan Clark},
  \bibinfo{person}{Diego de Las~Casas}, \bibinfo{person}{Aurelia Guy},
  \bibinfo{person}{Jacob Menick}, \bibinfo{person}{Roman Ring},
  \bibinfo{person}{Tom Hennigan}, \bibinfo{person}{Saffron Huang},
  \bibinfo{person}{Loren Maggiore}, \bibinfo{person}{Chris Jones},
  \bibinfo{person}{Albin Cassirer}, \bibinfo{person}{Andy Brock},
  \bibinfo{person}{Michela Paganini}, \bibinfo{person}{Geoffrey Irving},
  \bibinfo{person}{Oriol Vinyals}, \bibinfo{person}{Simon Osindero},
  \bibinfo{person}{Karen Simonyan}, \bibinfo{person}{Jack~W. Rae},
  \bibinfo{person}{Erich Elsen}, {and} \bibinfo{person}{Laurent Sifre}.}
  \bibinfo{year}{2021}\natexlab{}.
\newblock \showarticletitle{Improving Language Models by Retrieving from
  Trillions of Tokens}.
\newblock  (\bibinfo{year}{2021}).
\newblock
\showeprint[arxiv]{2112.04426}


\bibitem[Bowman et~al\mbox{.}(2015)]%
        {snli}
\bibfield{author}{\bibinfo{person}{Samuel~R. Bowman}, \bibinfo{person}{Gabor
  Angeli}, \bibinfo{person}{Christopher Potts}, {and}
  \bibinfo{person}{Christopher~D. Manning}.} \bibinfo{year}{2015}\natexlab{}.
\newblock \showarticletitle{A Large Annotated Corpus for Learning Natural
  Language Inference}. In \bibinfo{booktitle}{\emph{Proceedings of the 2015
  Conference on Empirical Methods in Natural Language Processing}}.
  \bibinfo{pages}{632--642}.
\newblock


\bibitem[Cer et~al\mbox{.}(2017)]%
        {stsb}
\bibfield{author}{\bibinfo{person}{Daniel Cer}, \bibinfo{person}{Mona Diab},
  \bibinfo{person}{Eneko Agirre}, \bibinfo{person}{I{\~n}igo Lopez-Gazpio},
  {and} \bibinfo{person}{Lucia Specia}.} \bibinfo{year}{2017}\natexlab{}.
\newblock \showarticletitle{{S}em{E}val-2017 Task 1: Semantic Textual
  Similarity Multilingual and Crosslingual Focused Evaluation}. In
  \bibinfo{booktitle}{\emph{Proceedings of the 11th International Workshop on
  Semantic Evaluation ({S}em{E}val-2017)}}.
\newblock


\bibitem[Chen et~al\mbox{.}(2020)]%
        {simclr}
\bibfield{author}{\bibinfo{person}{Ting Chen}, \bibinfo{person}{Simon
  Kornblith}, \bibinfo{person}{Mohammad Norouzi}, {and}
  \bibinfo{person}{Geoffrey Hinton}.} \bibinfo{year}{2020}\natexlab{}.
\newblock \showarticletitle{A Simple Framework for Contrastive Learning of
  Visual Representations}. In \bibinfo{booktitle}{\emph{Proceedings of the 37th
  International Conference on Machine Learning}}
  \emph{(\bibinfo{series}{Proceedings of Machine Learning Research},
  Vol.~\bibinfo{volume}{119})}. \bibinfo{pages}{1597--1607}.
\newblock


\bibitem[Chen et~al\mbox{.}(2022)]%
        {false_negative_detection}
\bibfield{author}{\bibinfo{person}{Tsai-Shien Chen}, \bibinfo{person}{Wei-Chih
  Hung}, \bibinfo{person}{Hung-Yu Tseng}, \bibinfo{person}{Shao-Yi Chien},
  {and} \bibinfo{person}{Ming-Hsuan Yang}.} \bibinfo{year}{2022}\natexlab{}.
\newblock \showarticletitle{Incremental False Negative Detection for
  Contrastive Learning}. In \bibinfo{booktitle}{\emph{International Conference
  on Learning Representations}}.
\newblock


\bibitem[Cheng et~al\mbox{.}(2020)]%
        {advaug}
\bibfield{author}{\bibinfo{person}{Yong Cheng}, \bibinfo{person}{Lu Jiang},
  \bibinfo{person}{Wolfgang Macherey}, {and} \bibinfo{person}{Jacob
  Eisenstein}.} \bibinfo{year}{2020}\natexlab{}.
\newblock \showarticletitle{{A}dv{A}ug: Robust Adversarial Augmentation for
  Neural Machine Translation}. In \bibinfo{booktitle}{\emph{Proceedings of the
  58th Annual Meeting of the Association for Computational Linguistics}}.
  \bibinfo{pages}{5961--5970}.
\newblock


\bibitem[Conneau and Kiela(2018)]%
        {senteval}
\bibfield{author}{\bibinfo{person}{Alexis Conneau} {and} \bibinfo{person}{Douwe
  Kiela}.} \bibinfo{year}{2018}\natexlab{}.
\newblock \showarticletitle{{S}ent{E}val: An Evaluation Toolkit for Universal
  Sentence Representations}. In \bibinfo{booktitle}{\emph{Proceedings of the
  Eleventh International Conference on Language Resources and Evaluation
  ({LREC} 2018)}}.
\newblock


\bibitem[Devlin et~al\mbox{.}(2019)]%
        {bert}
\bibfield{author}{\bibinfo{person}{Jacob Devlin}, \bibinfo{person}{Ming-Wei
  Chang}, \bibinfo{person}{Kenton Lee}, {and} \bibinfo{person}{Kristina
  Toutanova}.} \bibinfo{year}{2019}\natexlab{}.
\newblock \showarticletitle{{BERT: Pre-training of Deep Bidirectional
  Transformers for Language Understanding}}. In
  \bibinfo{booktitle}{\emph{Proceedings of the 2019 Conference of the North
  {A}merican Chapter of the Association for Computational Linguistics: Human
  Language Technologies, Volume 1 (Long and Short Papers)}}.
  \bibinfo{pages}{4171--4186}.
\newblock


\bibitem[Du et~al\mbox{.}(2021)]%
        {retrieval-self-train}
\bibfield{author}{\bibinfo{person}{Jingfei Du}, \bibinfo{person}{Edouard
  Grave}, \bibinfo{person}{Beliz Gunel}, \bibinfo{person}{Vishrav Chaudhary},
  \bibinfo{person}{Onur Celebi}, \bibinfo{person}{Michael Auli},
  \bibinfo{person}{Veselin Stoyanov}, {and} \bibinfo{person}{Alexis Conneau}.}
  \bibinfo{year}{2021}\natexlab{}.
\newblock \showarticletitle{Self-training Improves Pre-training for Natural
  Language Understanding}. In \bibinfo{booktitle}{\emph{Proceedings of the 2021
  Conference of the North American Chapter of the Association for Computational
  Linguistics: Human Language Technologies}}. \bibinfo{pages}{5408--5418}.
\newblock


\bibitem[Feng et~al\mbox{.}(2021)]%
        {survey-augmentation}
\bibfield{author}{\bibinfo{person}{Steven~Y. Feng}, \bibinfo{person}{Varun
  Gangal}, \bibinfo{person}{Jason Wei}, \bibinfo{person}{Sarath Chandar},
  \bibinfo{person}{Soroush Vosoughi}, \bibinfo{person}{Teruko Mitamura}, {and}
  \bibinfo{person}{Eduard Hovy}.} \bibinfo{year}{2021}\natexlab{}.
\newblock \showarticletitle{A Survey of Data Augmentation Approaches for
  {NLP}}. In \bibinfo{booktitle}{\emph{Findings of the Association for
  Computational Linguistics: ACL-IJCNLP 2021}}. \bibinfo{pages}{968--988}.
\newblock


\bibitem[Gao and Callan(2021a)]%
        {condenser}
\bibfield{author}{\bibinfo{person}{Luyu Gao} {and} \bibinfo{person}{Jamie
  Callan}.} \bibinfo{year}{2021}\natexlab{a}.
\newblock \showarticletitle{Condenser: a Pre-training Architecture for Dense
  Retrieval}. In \bibinfo{booktitle}{\emph{Proceedings of the 2021 Conference
  on Empirical Methods in Natural Language Processing}}.
  \bibinfo{pages}{981--993}.
\newblock


\bibitem[Gao and Callan(2021b)]%
        {cocondenser}
\bibfield{author}{\bibinfo{person}{Luyu Gao} {and} \bibinfo{person}{Jamie
  Callan}.} \bibinfo{year}{2021}\natexlab{b}.
\newblock \showarticletitle{{Unsupervised Corpus Aware Language Model
  Pre-training for Dense Passage Retrieval}}.
\newblock  (\bibinfo{year}{2021}).
\newblock
\showeprint[arxiv]{2108.05540}


\bibitem[Gao et~al\mbox{.}(2021)]%
        {simcse}
\bibfield{author}{\bibinfo{person}{Tianyu Gao}, \bibinfo{person}{Xingcheng
  Yao}, {and} \bibinfo{person}{Danqi Chen}.} \bibinfo{year}{2021}\natexlab{}.
\newblock \showarticletitle{{SimCSE: Simple Contrastive Learning of Sentence
  Embeddings}}. In \bibinfo{booktitle}{\emph{Proceedings of the 2021 Conference
  on Empirical Methods in Natural Language Processing}}.
  \bibinfo{pages}{6894--6910}.
\newblock


\bibitem[He et~al\mbox{.}(2020)]%
        {moco}
\bibfield{author}{\bibinfo{person}{Kaiming He}, \bibinfo{person}{Haoqi Fan},
  \bibinfo{person}{Yuxin Wu}, \bibinfo{person}{Saining Xie}, {and}
  \bibinfo{person}{Ross Girshick}.} \bibinfo{year}{2020}\natexlab{}.
\newblock \showarticletitle{Momentum Contrast for Unsupervised Visual
  Representation Learning}. In \bibinfo{booktitle}{\emph{2020 IEEE/CVF
  Conference on Computer Vision and Pattern Recognition (CVPR)}}.
  \bibinfo{pages}{9726--9735}.
\newblock


\bibitem[He et~al\mbox{.}(2021)]%
        {deberta}
\bibfield{author}{\bibinfo{person}{Pengcheng He}, \bibinfo{person}{Xiaodong
  Liu}, \bibinfo{person}{Jianfeng Gao}, {and} \bibinfo{person}{Weizhu Chen}.}
  \bibinfo{year}{2021}\natexlab{}.
\newblock \showarticletitle{{DeBERTa: Decoding-enhanced BERT with Disentangled
  Attention}}. In \bibinfo{booktitle}{\emph{International Conference on
  Learning Representations}}.
\newblock


\bibitem[Holtzman et~al\mbox{.}(2020)]%
        {nucleus-sampling}
\bibfield{author}{\bibinfo{person}{Ari Holtzman}, \bibinfo{person}{Jan Buys},
  \bibinfo{person}{Li Du}, \bibinfo{person}{Maxwell Forbes}, {and}
  \bibinfo{person}{Yejin Choi}.} \bibinfo{year}{2020}\natexlab{}.
\newblock \showarticletitle{The Curious Case of Neural Text Degeneration}. In
  \bibinfo{booktitle}{\emph{International Conference on Learning
  Representations}}.
\newblock


\bibitem[Jiang et~al\mbox{.}(2022)]%
        {prompt-bert}
\bibfield{author}{\bibinfo{person}{Ting Jiang}, \bibinfo{person}{Shaohan
  Huang}, \bibinfo{person}{Zihan Zhang}, \bibinfo{person}{Deqing Wang},
  \bibinfo{person}{Fuzhen Zhuang}, \bibinfo{person}{Furu Wei},
  \bibinfo{person}{Haizhen Huang}, \bibinfo{person}{Liangjie Zhang}, {and}
  \bibinfo{person}{Qi Zhang}.} \bibinfo{year}{2022}\natexlab{}.
\newblock \showarticletitle{{PromptBERT: Improving BERT Sentence Embeddings
  with Prompts}}.
\newblock  (\bibinfo{year}{2022}).
\newblock
\showeprint[arxiv]{2201.04337}


\bibitem[Johnson et~al\mbox{.}(2019)]%
        {faiss}
\bibfield{author}{\bibinfo{person}{Jeff Johnson}, \bibinfo{person}{Matthijs
  Douze}, {and} \bibinfo{person}{Herv{\'e} J{\'e}gou}.}
  \bibinfo{year}{2019}\natexlab{}.
\newblock \showarticletitle{Billion-scale Similarity Search with GPUs}.
\newblock \bibinfo{journal}{\emph{IEEE Transactions on Big Data}}
  \bibinfo{volume}{7}, \bibinfo{number}{3} (\bibinfo{year}{2019}),
  \bibinfo{pages}{535--547}.
\newblock


\bibitem[Kudo and Richardson(2018)]%
        {sentencepiece}
\bibfield{author}{\bibinfo{person}{Taku Kudo} {and} \bibinfo{person}{John
  Richardson}.} \bibinfo{year}{2018}\natexlab{}.
\newblock \showarticletitle{{S}entence{P}iece: A Simple and Language
  Independent Subword Tokenizer and Detokenizer for Neural Text Processing}. In
  \bibinfo{booktitle}{\emph{Proceedings of the 2018 Conference on Empirical
  Methods in Natural Language Processing: System Demonstrations}}.
  \bibinfo{pages}{66--71}.
\newblock


\bibitem[Li et~al\mbox{.}(2020b)]%
        {bert-flow}
\bibfield{author}{\bibinfo{person}{Bohan Li}, \bibinfo{person}{Hao Zhou},
  \bibinfo{person}{Junxian He}, \bibinfo{person}{Mingxuan Wang},
  \bibinfo{person}{Yiming Yang}, {and} \bibinfo{person}{Lei Li}.}
  \bibinfo{year}{2020}\natexlab{b}.
\newblock \showarticletitle{On the Sentence Embeddings from Pre-trained
  Language Models}. In \bibinfo{booktitle}{\emph{Proceedings of the 2020
  Conference on Empirical Methods in Natural Language Processing (EMNLP)}}.
  \bibinfo{pages}{9119--9130}.
\newblock


\bibitem[Li et~al\mbox{.}(2020a)]%
        {augmentation-lit}
\bibfield{author}{\bibinfo{person}{Chuanrong Li}, \bibinfo{person}{Lin
  Shengshuo}, \bibinfo{person}{Zeyu Liu}, \bibinfo{person}{Xinyi Wu},
  \bibinfo{person}{Xuhui Zhou}, {and} \bibinfo{person}{Shane
  Steinert-Threlkeld}.} \bibinfo{year}{2020}\natexlab{a}.
\newblock \showarticletitle{{Linguistically-Informed Transformations (LIT): A
  Method for Automatically Generating Contrast Sets}}. In
  \bibinfo{booktitle}{\emph{Proceedings of the Third BlackboxNLP Workshop on
  Analyzing and Interpreting Neural Networks for NLP}}.
  \bibinfo{pages}{126--135}.
\newblock


\bibitem[Liang et~al\mbox{.}(2021)]%
        {rdrop}
\bibfield{author}{\bibinfo{person}{Xiaobo Liang}, \bibinfo{person}{Lijun Wu},
  \bibinfo{person}{Juntao Li}, \bibinfo{person}{Yue Wang}, {and}
  \bibinfo{person}{Qi Meng}.} \bibinfo{year}{2021}\natexlab{}.
\newblock \showarticletitle{{R-Drop : Regularized Dropout for Neural
  Networks}}.
\newblock \bibinfo{journal}{\emph{Advances in Neural Information Processing
  Systems}}  \bibinfo{volume}{34} (\bibinfo{year}{2021}).
\newblock


\bibitem[Liu et~al\mbox{.}(2021)]%
        {survey-prompt}
\bibfield{author}{\bibinfo{person}{Pengfei Liu}, \bibinfo{person}{Weizhe Yuan},
  \bibinfo{person}{Jinlan Fu}, \bibinfo{person}{Zhengbao Jiang},
  \bibinfo{person}{Hiroaki Hayashi}, {and} \bibinfo{person}{Graham Neubig}.}
  \bibinfo{year}{2021}\natexlab{}.
\newblock \showarticletitle{{Pre-train, Prompt, and Predict: A Systematic
  Survey of Prompting Methods in Natural Language Processing}}.
\newblock  (\bibinfo{year}{2021}).
\newblock
\showeprint[arxiv]{2107.13586}


\bibitem[Liu et~al\mbox{.}(2020)]%
        {alum}
\bibfield{author}{\bibinfo{person}{Xiaodong Liu}, \bibinfo{person}{Hao Cheng},
  \bibinfo{person}{Pengcheng He}, \bibinfo{person}{Weizhu Chen},
  \bibinfo{person}{Yu Wang}, \bibinfo{person}{Hoifung Poon}, {and}
  \bibinfo{person}{Jianfeng Gao}.} \bibinfo{year}{2020}\natexlab{}.
\newblock \showarticletitle{{Adversarial Training for Large Neural Language
  Models}}.
\newblock  (\bibinfo{year}{2020}).
\newblock
\showeprint[arxiv]{2004.08994}


\bibitem[Liu et~al\mbox{.}(2019)]%
        {roberta}
\bibfield{author}{\bibinfo{person}{Yinhan Liu}, \bibinfo{person}{Myle Ott},
  \bibinfo{person}{Naman Goyal}, \bibinfo{person}{Jingfei Du},
  \bibinfo{person}{Mandar Joshi}, \bibinfo{person}{Danqi Chen},
  \bibinfo{person}{Omer Levy}, \bibinfo{person}{Mike Lewis},
  \bibinfo{person}{Luke Zettlemoyer}, {and} \bibinfo{person}{Veselin
  Stoyanov}.} \bibinfo{year}{2019}\natexlab{}.
\newblock \showarticletitle{{RoBERTa: A Robustly Optimized BERT Pretraining
  Approach}}.
\newblock  (\bibinfo{year}{2019}).
\newblock
\showeprint[arxiv]{1907.11692}


\bibitem[Meng et~al\mbox{.}(2021)]%
        {cocolm}
\bibfield{author}{\bibinfo{person}{Yu Meng}, \bibinfo{person}{Chenyan Xiong},
  \bibinfo{person}{Payal Bajaj}, \bibinfo{person}{Saurabh Tiwary},
  \bibinfo{person}{Paul Bennett}, \bibinfo{person}{Jiawei Han}, {and}
  \bibinfo{person}{Xia Song}.} \bibinfo{year}{2021}\natexlab{}.
\newblock \showarticletitle{{COCO-LM: Correcting and Contrasting Text Sequences
  for Language Model Pretraining}}.
\newblock \bibinfo{journal}{\emph{Advances in Neural Information Processing
  Systems}}  \bibinfo{volume}{34} (\bibinfo{year}{2021}).
\newblock


\bibitem[Min et~al\mbox{.}(2021)]%
        {survey}
\bibfield{author}{\bibinfo{person}{Bonan Min}, \bibinfo{person}{Hayley Ross},
  \bibinfo{person}{Elior Sulem}, \bibinfo{person}{Amir Pouran~Ben Veyseh},
  \bibinfo{person}{Thien~Huu Nguyen}, \bibinfo{person}{Oscar Sainz},
  \bibinfo{person}{Eneko Agirre}, \bibinfo{person}{Ilana Heinz}, {and}
  \bibinfo{person}{Dan Roth}.} \bibinfo{year}{2021}\natexlab{}.
\newblock \showarticletitle{{Recent Advances in Natural Language Processing via
  Large Pre-Trained Language Models: A Survey}}.
\newblock  (\bibinfo{year}{2021}).
\newblock
\showeprint[arxiv]{2111.01243}


\bibitem[Min et~al\mbox{.}(2020)]%
        {augmentation-syntactic}
\bibfield{author}{\bibinfo{person}{Junghyun Min}, \bibinfo{person}{R.~Thomas
  McCoy}, \bibinfo{person}{Dipanjan Das}, \bibinfo{person}{Emily Pitler}, {and}
  \bibinfo{person}{Tal Linzen}.} \bibinfo{year}{2020}\natexlab{}.
\newblock \showarticletitle{{Syntactic Data Augmentation Increases Robustness
  to Inference Heuristics}}. In \bibinfo{booktitle}{\emph{Proceedings of the
  58th Annual Meeting of the Association for Computational Linguistics}}.
  \bibinfo{pages}{2339--2352}.
\newblock


\bibitem[Mosbach et~al\mbox{.}(2020)]%
        {finetune-bert}
\bibfield{author}{\bibinfo{person}{Marius Mosbach}, \bibinfo{person}{Maksym
  Andriushchenko}, {and} \bibinfo{person}{Dietrich Klakow}.}
  \bibinfo{year}{2020}\natexlab{}.
\newblock \showarticletitle{{On the Stability of Fine-tuning BERT:
  Misconceptions, Explanations, and Strong Baselines}}. In
  \bibinfo{booktitle}{\emph{International Conference on Learning
  Representations}}.
\newblock


\bibitem[Provilkov et~al\mbox{.}(2020)]%
        {bpe-dropout}
\bibfield{author}{\bibinfo{person}{Ivan Provilkov}, \bibinfo{person}{Dmitrii
  Emelianenko}, {and} \bibinfo{person}{Elena Voita}.}
  \bibinfo{year}{2020}\natexlab{}.
\newblock \showarticletitle{{BPE}-Dropout: Simple and Effective Subword
  Regularization}. In \bibinfo{booktitle}{\emph{Proceedings of the 58th Annual
  Meeting of the Association for Computational Linguistics}}.
  \bibinfo{pages}{1882--1892}.
\newblock


\bibitem[Qu et~al\mbox{.}(2020)]%
        {coda}
\bibfield{author}{\bibinfo{person}{Yanru Qu}, \bibinfo{person}{Dinghan Shen},
  \bibinfo{person}{Yelong Shen}, \bibinfo{person}{Sandra Sajeev},
  \bibinfo{person}{Jiawei Han}, {and} \bibinfo{person}{Weizhu Chen}.}
  \bibinfo{year}{2020}\natexlab{}.
\newblock \showarticletitle{{CoDA: Contrast-enhanced and Diversity-promoting
  Data Augmentation for Natural Language Understanding}}. In
  \bibinfo{booktitle}{\emph{International Conference on Learning
  Representations}}.
\newblock


\bibitem[Radford et~al\mbox{.}(2019)]%
        {gpt2}
\bibfield{author}{\bibinfo{person}{Alec Radford}, \bibinfo{person}{Jeffrey Wu},
  \bibinfo{person}{Rewon Child}, \bibinfo{person}{David Luan},
  \bibinfo{person}{Dario Amodei}, \bibinfo{person}{Ilya Sutskever},
  {et~al\mbox{.}}} \bibinfo{year}{2019}\natexlab{}.
\newblock \showarticletitle{Language models are unsupervised multitask
  learners}.
\newblock \bibinfo{journal}{\emph{OpenAI blog}} (\bibinfo{year}{2019}).
\newblock


\bibitem[Robinson et~al\mbox{.}(2021)]%
        {cl_sample_hard_negatives}
\bibfield{author}{\bibinfo{person}{Joshua~David Robinson},
  \bibinfo{person}{Ching-Yao Chuang}, \bibinfo{person}{Suvrit Sra}, {and}
  \bibinfo{person}{Stefanie Jegelka}.} \bibinfo{year}{2021}\natexlab{}.
\newblock \showarticletitle{Contrastive Learning with Hard Negative Samples}.
  In \bibinfo{booktitle}{\emph{International Conference on Learning
  Representations}}.
\newblock


\bibitem[Sennrich et~al\mbox{.}(2016a)]%
        {back-translation}
\bibfield{author}{\bibinfo{person}{Rico Sennrich}, \bibinfo{person}{Barry
  Haddow}, {and} \bibinfo{person}{Alexandra Birch}.}
  \bibinfo{year}{2016}\natexlab{a}.
\newblock \showarticletitle{Improving Neural Machine Translation Models with
  Monolingual Data}. In \bibinfo{booktitle}{\emph{Proceedings of the 54th
  Annual Meeting of the Association for Computational Linguistics (Volume 1:
  Long Papers)}}. \bibinfo{pages}{86--96}.
\newblock


\bibitem[Sennrich et~al\mbox{.}(2016b)]%
        {bpe}
\bibfield{author}{\bibinfo{person}{Rico Sennrich}, \bibinfo{person}{Barry
  Haddow}, {and} \bibinfo{person}{Alexandra Birch}.}
  \bibinfo{year}{2016}\natexlab{b}.
\newblock \showarticletitle{Neural Machine Translation of Rare Words with
  Subword Units}. In \bibinfo{booktitle}{\emph{Proceedings of the 54th Annual
  Meeting of the Association for Computational Linguistics (Volume 1: Long
  Papers)}}. \bibinfo{pages}{1715--1725}.
\newblock


\bibitem[Su et~al\mbox{.}(2021)]%
        {bert-whitening}
\bibfield{author}{\bibinfo{person}{Jianlin Su}, \bibinfo{person}{Jiarun Cao},
  \bibinfo{person}{Weijie Liu}, {and} \bibinfo{person}{Yangyiwen Ou}.}
  \bibinfo{year}{2021}\natexlab{}.
\newblock \showarticletitle{{Whitening Sentence Representations for Better
  Semantics and Faster Retrieval}}.
\newblock  (\bibinfo{year}{2021}).
\newblock
\showeprint[arxiv]{2103.15316}


\bibitem[Wang et~al\mbox{.}(2019)]%
        {glue}
\bibfield{author}{\bibinfo{person}{Alex Wang}, \bibinfo{person}{Amanpreet
  Singh}, \bibinfo{person}{Julian Michael}, \bibinfo{person}{Felix Hill},
  \bibinfo{person}{Omer Levy}, {and} \bibinfo{person}{Samuel~R. Bowman}.}
  \bibinfo{year}{2019}\natexlab{}.
\newblock \showarticletitle{{GLUE: A Multi-task Benchmark and Analysis Platform
  for Natural Language Understanding}}. In
  \bibinfo{booktitle}{\emph{International Conference on Learning
  Representations}}.
\newblock


\bibitem[Wang and Liu(2021)]%
        {cl_temperature}
\bibfield{author}{\bibinfo{person}{Feng Wang} {and} \bibinfo{person}{Huaping
  Liu}.} \bibinfo{year}{2021}\natexlab{}.
\newblock \showarticletitle{Understanding the Behaviour of Contrastive Loss}.
  In \bibinfo{booktitle}{\emph{Proceedings of the IEEE/CVF Conference on
  Computer Vision and Pattern Recognition (CVPR)}}.
  \bibinfo{pages}{2495--2504}.
\newblock


\bibitem[Wei and Zou(2019)]%
        {eda}
\bibfield{author}{\bibinfo{person}{Jason Wei} {and} \bibinfo{person}{Kai Zou}.}
  \bibinfo{year}{2019}\natexlab{}.
\newblock \showarticletitle{EDA: Easy Data Augmentation Techniques for Boosting
  Performance on Text Classification Tasks}. In
  \bibinfo{booktitle}{\emph{Proceedings of the 2019 Conference on Empirical
  Methods in Natural Language Processing and the 9th International Joint
  Conference on Natural Language Processing (EMNLP-IJCNLP)}}.
  \bibinfo{pages}{6382--6388}.
\newblock


\bibitem[Welleck et~al\mbox{.}(2019)]%
        {unlikelihood}
\bibfield{author}{\bibinfo{person}{Sean Welleck}, \bibinfo{person}{Ilia
  Kulikov}, \bibinfo{person}{Stephen Roller}, \bibinfo{person}{Emily Dinan},
  \bibinfo{person}{Kyunghyun Cho}, {and} \bibinfo{person}{Jason Weston}.}
  \bibinfo{year}{2019}\natexlab{}.
\newblock \showarticletitle{{Neural Text Generation with Unlikelihood
  Training}}. In \bibinfo{booktitle}{\emph{International Conference on Learning
  Representations}}.
\newblock


\bibitem[Williams et~al\mbox{.}(2018)]%
        {mnli}
\bibfield{author}{\bibinfo{person}{Adina Williams}, \bibinfo{person}{Nikita
  Nangia}, {and} \bibinfo{person}{Samuel Bowman}.}
  \bibinfo{year}{2018}\natexlab{}.
\newblock \showarticletitle{A Broad-Coverage Challenge Corpus for Sentence
  Understanding through Inference}. In \bibinfo{booktitle}{\emph{Proceedings of
  the 2018 Conference of the North {A}merican Chapter of the Association for
  Computational Linguistics: Human Language Technologies, Volume 1 (Long
  Papers)}}. \bibinfo{pages}{1112--1122}.
\newblock


\bibitem[Wolf et~al\mbox{.}(2020)]%
        {huggingface}
\bibfield{author}{\bibinfo{person}{Thomas Wolf}, \bibinfo{person}{Lysandre
  Debut}, \bibinfo{person}{Victor Sanh}, \bibinfo{person}{Julien Chaumond},
  \bibinfo{person}{Clement Delangue}, \bibinfo{person}{Anthony Moi},
  \bibinfo{person}{Pierric Cistac}, \bibinfo{person}{Tim Rault},
  \bibinfo{person}{Rémi Louf}, \bibinfo{person}{Morgan Funtowicz},
  \bibinfo{person}{Joe Davison}, \bibinfo{person}{Sam Shleifer},
  \bibinfo{person}{Patrick von Platen}, \bibinfo{person}{Clara Ma},
  \bibinfo{person}{Yacine Jernite}, \bibinfo{person}{Julien Plu},
  \bibinfo{person}{Canwen Xu}, \bibinfo{person}{Teven~Le Scao},
  \bibinfo{person}{Sylvain Gugger}, \bibinfo{person}{Mariama Drame},
  \bibinfo{person}{Quentin Lhoest}, {and} \bibinfo{person}{Alexander~M. Rush}.}
  \bibinfo{year}{2020}\natexlab{}.
\newblock \showarticletitle{HuggingFace's Transformers: State-of-the-Art
  Natural Language Processing}. In \bibinfo{booktitle}{\emph{Proceedings of the
  2020 Conference on Empirical Methods in Natural Language Processing: System
  Demonstrations}}. \bibinfo{pages}{38--45}.
\newblock


\bibitem[Wu et~al\mbox{.}(2021)]%
        {esimcse}
\bibfield{author}{\bibinfo{person}{Xing Wu}, \bibinfo{person}{Chaochen Gao},
  \bibinfo{person}{Liangjun Zang}, \bibinfo{person}{Jizhong Han},
  \bibinfo{person}{Zhongyuan Wang}, {and} \bibinfo{person}{Songlin Hu}.}
  \bibinfo{year}{2021}\natexlab{}.
\newblock \showarticletitle{{ESimCSE: Enhanced Sample Building Method for
  Contrastive Learning of Unsupervised Sentence Embedding}}.
\newblock  (\bibinfo{year}{2021}).
\newblock
\showeprint[arxiv]{2109.04380}


\bibitem[Xiong et~al\mbox{.}(2021)]%
        {retrieval-approximate-nn}
\bibfield{author}{\bibinfo{person}{Lee Xiong}, \bibinfo{person}{Chenyan Xiong},
  \bibinfo{person}{Ye Li}, \bibinfo{person}{Kwok-Fung Tang},
  \bibinfo{person}{Jialin Liu}, \bibinfo{person}{Paul~N. Bennett},
  \bibinfo{person}{Junaid Ahmed}, {and} \bibinfo{person}{Arnold Overwijk}.}
  \bibinfo{year}{2021}\natexlab{}.
\newblock \showarticletitle{Approximate Nearest Neighbor Negative Contrastive
  Learning for Dense Text Retrieval}. In
  \bibinfo{booktitle}{\emph{International Conference on Learning
  Representations}}.
\newblock


\bibitem[Zhan et~al\mbox{.}(2021)]%
        {retrieval_dynamic_hard_negatives}
\bibfield{author}{\bibinfo{person}{Jingtao Zhan}, \bibinfo{person}{Jiaxin Mao},
  \bibinfo{person}{Yiqun Liu}, \bibinfo{person}{Jiafeng Guo},
  \bibinfo{person}{Min Zhang}, {and} \bibinfo{person}{Shaoping Ma}.}
  \bibinfo{year}{2021}\natexlab{}.
\newblock \showarticletitle{Optimizing Dense Retrieval Model Training with Hard
  Negatives}. In \bibinfo{booktitle}{\emph{Proceedings of the 44th
  International ACM SIGIR Conference on Research and Development in Information
  Retrieval}}. \bibinfo{pages}{1503–1512}.
\newblock


\bibitem[Zhang et~al\mbox{.}(2021)]%
        {vascl}
\bibfield{author}{\bibinfo{person}{Dejiao Zhang}, \bibinfo{person}{Wei Xiao},
  \bibinfo{person}{Henghui Zhu}, \bibinfo{person}{Xiaofei Ma}, {and}
  \bibinfo{person}{Andrew~O. Arnold}.} \bibinfo{year}{2021}\natexlab{}.
\newblock \showarticletitle{{Virtual Augmentation Supported Contrastive
  Learning of Sentence Representations}}.
\newblock  (\bibinfo{year}{2021}).
\newblock
\showeprint[arxiv]{2110.08552}


\bibitem[Zhang et~al\mbox{.}(2019)]%
        {yopo}
\bibfield{author}{\bibinfo{person}{Dinghuai Zhang}, \bibinfo{person}{Tianyuan
  Zhang}, \bibinfo{person}{Yiping Lu}, \bibinfo{person}{Zhanxing Zhu}, {and}
  \bibinfo{person}{Bin Dong}.} \bibinfo{year}{2019}\natexlab{}.
\newblock \showarticletitle{You Only Propagate Once: Accelerating Adversarial
  Training via Maximal Principle}. In \bibinfo{booktitle}{\emph{Advances in
  Neural Information Processing Systems}}. \bibinfo{pages}{227--238}.
\newblock


\bibitem[Zhou et~al\mbox{.}(2021)]%
        {freq_bias_context_embds}
\bibfield{author}{\bibinfo{person}{Kaitlyn Zhou}, \bibinfo{person}{Kawin
  Ethayarajh}, {and} \bibinfo{person}{Dan Jurafsky}.}
  \bibinfo{year}{2021}\natexlab{}.
\newblock \showarticletitle{{Frequency-based Distortions in Contextualized Word
  Embeddings}}.
\newblock  (\bibinfo{year}{2021}).
\newblock
\showeprint[arxiv]{2104.08465}


\bibitem[Zhu et~al\mbox{.}(2020)]%
        {freelb}
\bibfield{author}{\bibinfo{person}{Chen Zhu}, \bibinfo{person}{Yu Cheng},
  \bibinfo{person}{Zhe Gan}, \bibinfo{person}{Siqi Sun}, \bibinfo{person}{Tom
  Goldstein}, {and} \bibinfo{person}{Jingjing Liu}.}
  \bibinfo{year}{2020}\natexlab{}.
\newblock \showarticletitle{FreeLB: Enhanced Adversarial Training for Natural
  Language Understanding}. In \bibinfo{booktitle}{\emph{International
  Conference on Learning Representations}}.
\newblock


\end{thebibliography}

% \clearpage
\begin{appendices}

\begin{table*}[ht]
	\caption{Performance of \texorpdfstring{DeBERTa\textsubscript{1.5B}}{DeBERTa-1.5B} with R-drop and switch-case on GLUE development set.}
	\centering
	\label{Tab:glue_results}
	\begin{threeparttable}
		\begin{tabular}{l c c c c c c c c c}
			\toprule
			Task & MNLI-m & QQP & QNLI & SST-2 & CoLA & STS-B & MRPC & RTE & Avg. \\
			Dataset size & 392k & 363k & 108k & 67k & 8.5k & 5.7k & 3.5k & 2.5k & - \\
			\midrule
			\texorpdfstring{DeBERTa\textsubscript{1.5B}}{DeBERTa-1.5B} & 91.7 & 92.7 & 96.0 & \textbf{97.2} & 72.0 & 92.9 & 92.0 & 93.5 & 90.7 \\
			\(\;\;+\) R-drop & \textbf{91.9} & \textbf{92.9} & 96.1 & \textbf{97.2} & 73.8 & 93.2 & 93.1 & 93.7 & 91.2 \\
			\(\;\;+\) R-drop \(+\) switch-case & \textbf{92.0} & \textbf{93.0} & \textbf{96.3} & \textbf{97.2} & \textbf{75.5} & \textbf{93.6} & \textbf{93.9} & \textbf{94.2} & \textbf{91.7} \\
			\bottomrule
		\end{tabular}
		\iffalse
		\begin{tablenotes}
			\item[1] 
		\end{tablenotes}
		\fi
	\end{threeparttable}
\end{table*}

\section{Switch-case experiments on GLUE}
\label{appendix:glue}

\subsection{Evaluation Setup}
To show switch-case is a general data augmentation approach applicable beyond RoBERTa models and sentence embedding learning tasks, we evaluate it on \texorpdfstring{DeBERTa\textsubscript{1.5B}}{DeBERTa-1.5B} \cite{deberta} and the General Language Understanding Evaluation (GLUE) benchmark \cite{glue}. 

\paragraph{Datasets} GLUE contains eight tasks covering natural language inference (MNLI, RTE and QNLI), semantic similarity (MRPC, QQP and STS-B), sentiment classification (SST-2), and linguistic acceptability classification (CoLA). The evaluation metrics are Matthews correlation for CoLA, Spearman correlation for STS-B, the MNLI-match accuracy for MNLI, and accuracy for the rest. We also report the average of metrics of these eight tasks as the overall performance standard.

\paragraph{Baselines} We follow the framework of R-drop \cite{rdrop}, which shares a similar idea with SimCSE. R-drop passes the same batch to the encoder twice with different dropout masks, and calculates the distance of the two output representations using the symmetric KL-divergence. The distance is used as a regularization loss added to the original task objective. We apply switch-case augmentation to one side of R-drop, i.e., dropout to create one view, dropout+switch-case for the other, and compare it against the R-drop baseline.

\paragraph{Implementation Details} We adopt the pretrained \texorpdfstring{DeBERTa\textsubscript{1.5B}}{DeBERTa-1.5B} model as the encoder. With 1.5 billion parameters, it surpasses a majority of pre-trained models on GLUE benchmark with a large margin. A simple linear classification or regression head is placed on top of the encoder to handle each GLUE task, and the overall model is trained in a supervised manner. We mainly follow the hyper-parameter settings of the original DeBERTa paper \cite{deberta}, except that we train the model for 20 epochs on small tasks (CoLA, MRPC, STS-B and RTE) to alleviate the large variance \cite{finetune-bert}, and 4 epochs for the rest tasks. We select the weight of the R-drop KL-divergence loss in \(\{1,2,3,4\}\), and switch-case probability \(p_{sc}\) in \(\{0.05,0.1,0.15\}\). All GLUE scores are reported based on the development dataset.

\subsection{results}
As shown in Tab. \ref{Tab:glue_results}, switch-case can be used as a data augmentation method to further improve R-drop's performance on natural language understanding tasks. Specifically, when applied on \texorpdfstring{DeBERTa\textsubscript{1.5B}}{DeBERTa-1.5B}, R-drop improves the average GLUE score from baseline's 90.7 to 91.2, while the proposed switch-case further increases it to 91.7. As with many data augmentation methods (including dropout in R-drop), switch-case is most effective on small tasks, eps. CoLA and MRPC. Switch-case is not effective on SST-2, perhaps because sentiment analysis relies more on the understanding of a few keywords than the whole sentence. We also note that \begin{enumerate*}
  \item \texorpdfstring{DeBERTa\textsubscript{1.5B}}{DeBERTa-1.5B} uses Sentencepiece tokenizer \cite{sentencepiece} with a 128k vocabulary, rather than RoBERTa's BPE tokenizer \cite{bpe} with 50k vocabulary, showing that switch-case works across different tokenizer settings.
  \item \texorpdfstring{DeBERTa\textsubscript{1.5B}}{DeBERTa-1.5B} has a much larger model size (1.5B) than \texorpdfstring{RoBERTa\textsubscript{large}}{RoBERTa-large} (355M), showing that the bias of pretrained word embeddings still exists in larger models.
\end{enumerate*}

\end{appendices}

\end{document}